\documentclass[10pt,twocolumn,letterpaper]{article}

\usepackage[pagenumbers]{cvpr} % To force page numbers, e.g. for an arXiv version

\usepackage{amssymb}% http://ctan.org/pkg/amssymb
\usepackage{pifont}% http://ctan.org/pkg/pifont

\usepackage[dvipsnames]{xcolor}

\newcommand{\envelope}{\ding{41}}

\definecolor{cvprblue}{rgb}{0.21,0.49,0.74}
\usepackage[pagebackref,breaklinks,colorlinks,citecolor=cvprblue]{hyperref}
\usepackage{multirow}
\usepackage[T1]{fontenc}

\usepackage{diagbox}   % 提供斜线表头命令
\usepackage{algorithm}
\usepackage{algorithmic}
\usepackage{multirow}
\usepackage{amsmath}
\usepackage{amssymb}
\usepackage{graphicx}
\usepackage[table]{xcolor}

\title{Light-Weight Cross-Modal Enhancement Method with Benchmark Construction for UAV-based Open-Vocabulary Object Detection}

\author{
Zhenhai Weng$^{1}$,~
Xinjie Li$^{1}$,~
Can Wu$^{1}$,~
Weijie He$^{1}$,~
Jianfeng Lv$^{2}$,~
Dong Zhou$^{3,4}$,~
Zhongliang Yu$^{1,\text{\envelope}}$,~\\
\textbf{\small $^\text{\envelope}$ corresponding author}\\
$^1$~School of Automation, Chongqing University\\
$^2$~School of Information Science and Engineering, Lanzhou University\\
$^3$~Department of Control Science and Engineering, Harbin Institute of Technology\\
$^4$~Department of Mechanical and Automation Engineering, The Chinese University of Hong Kong\\
}

\begin{document}
\maketitle

\begin{abstract}
Open-Vocabulary Object Detection (OVD) faces severe performance degradation when applied to UAV imagery due to the domain gap from ground-level datasets. To address this challenge, we propose a complete UAV-oriented solution that combines both dataset construction and model innovation. First, we design a refined UAV-Label Engine, which efficiently resolves annotation redundancy, inconsistency, and ambiguity, enabling the generation of large-scale UAV datasets. Based on this engine, we construct two new benchmarks: UAVDE-2M, with over 2.4M instances across 1,800+ categories, and UAVCAP-15K, providing rich image-text pairs for vision-language pretraining. Second, we introduce the Cross-Attention Gated Enhancement (CAGE) module, a lightweight dual-path fusion design that integrates cross-attention, adaptive gating, and global FiLM modulation for robust text–vision alignment. By embedding CAGE into the YOLO-World-v2 framework, our method achieves significant gains in both accuracy and efficiency, notably improving zero-shot detection on VisDrone by +5.3 mAP while reducing parameters and GFLOPs, and demonstrating strong cross-domain generalization on SIMD. Extensive experiments and real-world UAV deployment confirm the effectiveness and practicality of our proposed solution for UAV-based OVD.
\end{abstract}

\section{Introduction}
\label{sec:intro}
Object detection, a fundamental task in computer vision, has witnessed rapid progress in recent years. However, traditional object detectors are limited to performing well only on object categories seen during training. This closed-set detection paradigm severely limits their application in open-world scenarios, particularly for agents like robots or UAVs. To address this limitation, researchers have introduced open-vocabulary object detection (OVD)\cite{9577418, liu2023grounding}. By leveraging vision-language models (VLMs) such as CLIP\cite{clip_ICML} to align textual and visual information, OVD methods can detect novel categories not present in the training set, demonstrating remarkable zero-shot capabilities. Nevertheless, prior OVD models often rely on computationally intensive backbones or detectors with a large number of parameters, making them difficult to deploy on resource-constrained mobile platforms like robots and UAVs. In response to this challenge, real-time OVD methods have recently emerged\cite{wang2025yoloerealtimeseeing, Cheng2024YOLOWorld}. These methods strike a balance between high detection accuracy and real-time performance, enabling deployment on edge computing devices.\par
\begin{figure}[t]
	\centering
	\includegraphics[width=0.45\textwidth]{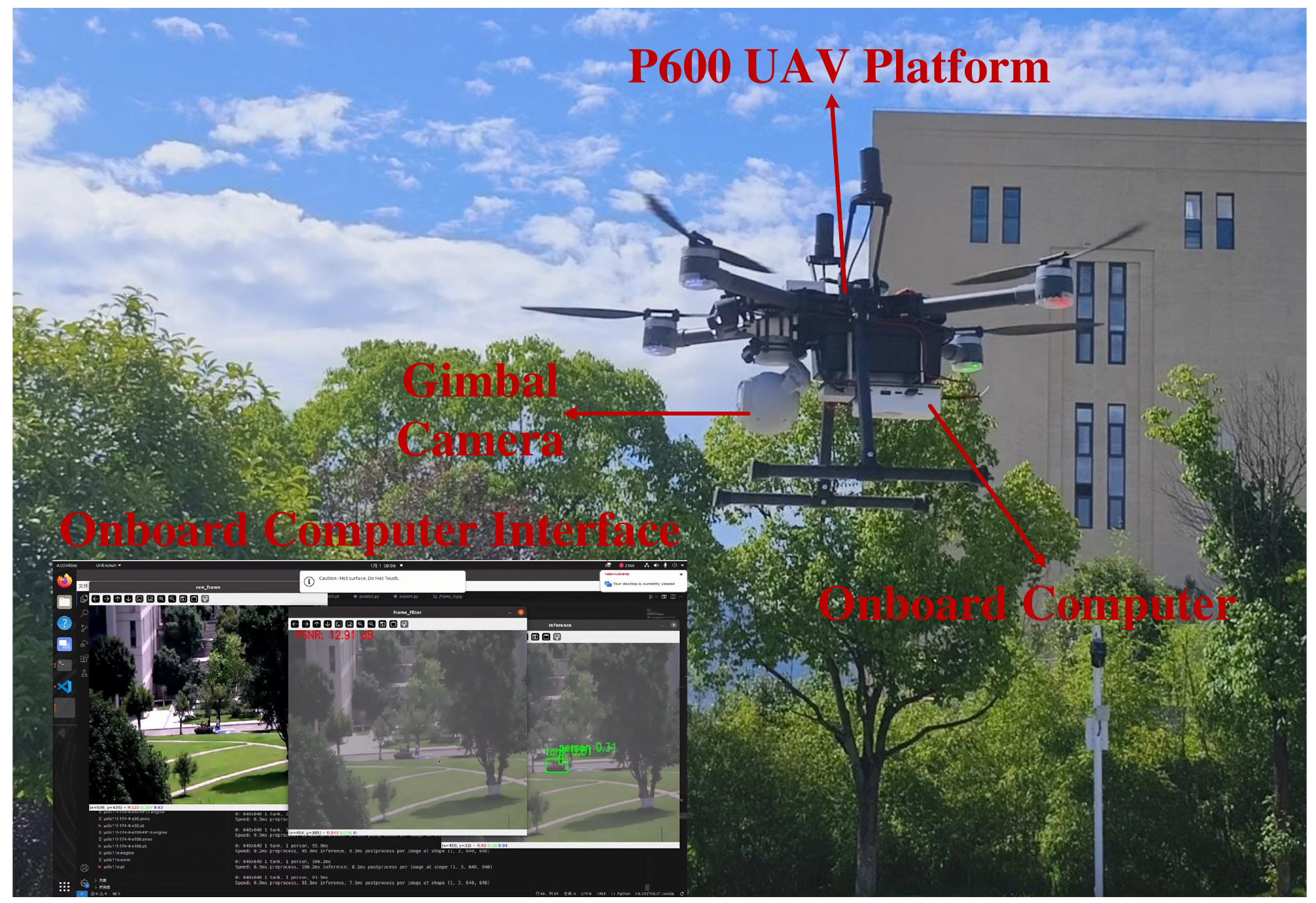}
	\caption{\textbf{Image of our UAV platform.} Our UAV experimental platform is equipped with a gimbal camera and an onboard computing device (NVIDIA ORIN NX 16G).}
	\label{fig:P600}
\end{figure}
\begin{figure}[htbp]
	\centering
	\includegraphics[width=0.45\textwidth]{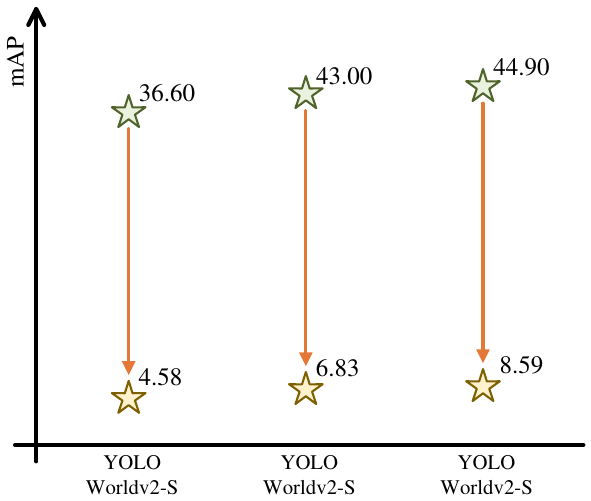}
	\caption{\textbf{Comparison of YOLO-World's Zero-Shot Performance on the COCO and VisDrone Datasets.} The \textcolor[RGB]{80,99,42}{green} stars demonstrate the mAP in COCO dataset, and the \textcolor[RGB]{127,96,0}{yellow} ones demonstrate the mAP in VosDrone dataset. Specifically, for the VisDrone dataset, we use a confidence threshold of 0.001, a setting that is kept consistent across all following experiments.}
	\label{fig:performances between natural-world images and drone perspective images}
\end{figure}
\begin{figure}[htbp]
	\centering
	\includegraphics[width=0.45\textwidth]{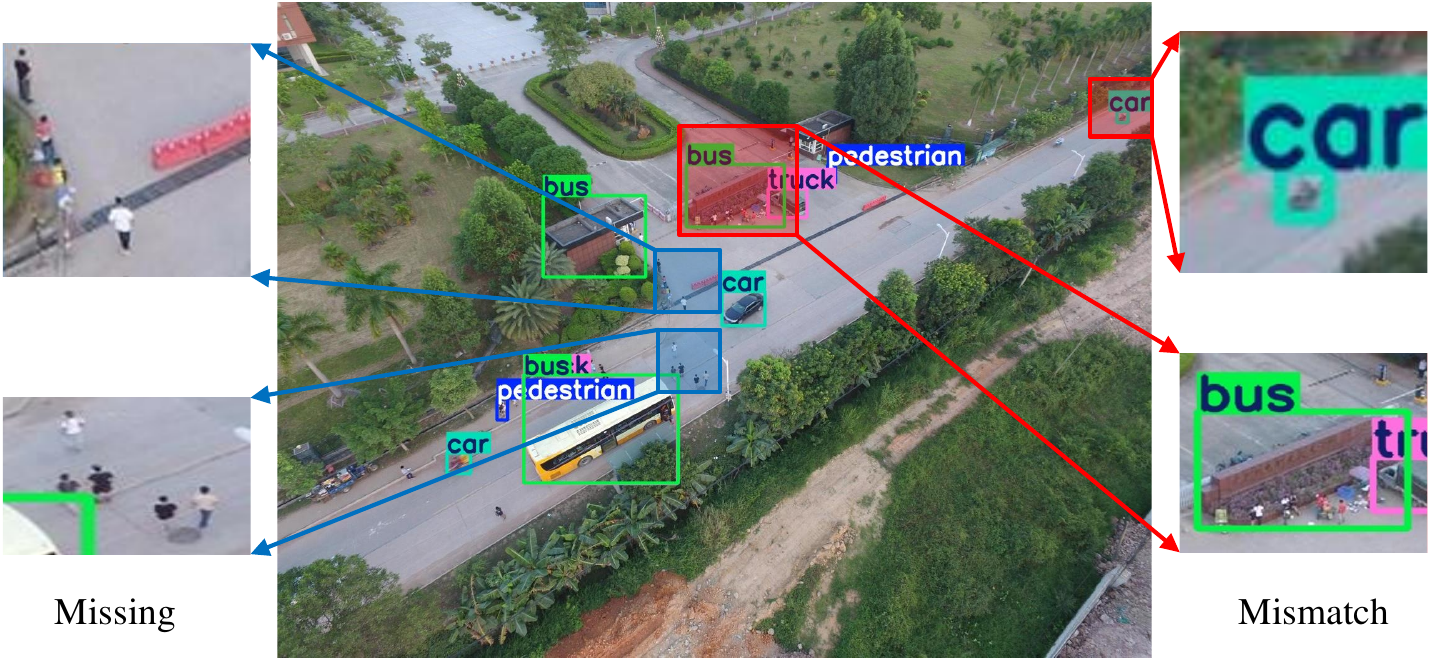}
	\caption{\textbf{Common Challenges in Open-Vocabulary Detection for UAV Imagery.} Common failure modes for OVD in aerial perspectives include false negatives (missed detections) and semantic mismatches.}
	\label{fig:missing and mismatch}
\end{figure}

However, although existing real-time OVD methods perform well on natural-world images datasets(e.g. LVIS\cite{gupta2019lvis}, COCO\cite{lin2015microsoft}), their performance on UAV-view object detection benchmarks(e.g. VisDrone\cite{visdrone}, DroneVehicle\cite{sun2020drone}) is relatively poor, shown in Figure \ref{fig:performances between natural-world images and drone perspective images} and Figure \ref{fig:missing and mismatch}. Several factors likely account for this discrepancy. First, compared with conventional natural-image scenes, UAV imagery generally contains more cluttered backgrounds and objects that occupy far fewer pixels. Moreover, current open-vocabulary object detectors are trained almost exclusively on large-scale natural-world images datasets(e.g. Objects365\cite{shao2019objects365}, GoldG\cite{GoldG}), and this domain mismatch between training data and UAV views results in the pronounced drop in performance. \par
Therefore, in this paper we are motivated to advance real-time OVD for UAV perspective images. To achieve this goal, first, we optimized the LAE-Label Engine\cite{pan2025locate} and proposed the UAV-Label Engine. The UAV-Label Engine improves annotation accuracy and not only generates object detection datasets but also produces caption datasets, which are widely used in open-vocabulary detection tasks. Next, we used the UAV-Label Engine to generate our UAVDE-2M dataset and UAVCAP-15K dataset. UAVDE-2M is a huge UAV's view object detection dataset with over 3 million instance annotations and more than 1,800 object categories. UAVCAP-15K is a caption dataset containing over 10,000 image-text pairs. Both datasets' images are sourced from publicly available UAV-view task datasets.\par
To effectively embed textual guidance into UAV visual representations, we propose the cross-attention gated enhancement(CAGE) fusion module. The module's innovation lies in a dual-path architecture that leverages a cross-attention governed by an gate mechanism to ensure precise, spatially-aware semantic grounding, while a parallel FiLM layer provides global feature modulation, with both pathways integrated into the visual stream via a residual connection for robust enhancement.\par
Finally, to validate the practical efficacy of our model and dataset in real-world UAV applications, we deployed the trained model onto our custom UAV experimental platform and conducted a series of object detection experiments.We Figure \ref{fig:P600} illustrates the UAV platform.\par
We summarize the main contributions as follows,
\begin{itemize}
	\item We introduce two large-scale UAV-view datasets, UAVDE-2M and UAVCAP-15K, to facilitate the pre-training of open-vocabulary object detection models.
	\item We propose CAGE, a lightweight dual-path gated cross-attention fusion module, and integrate it into the YOLO-Worldv2 model.
	\item We conduct comparative experiments on the VisDrone dataset and in real-world UAV scenarios, demonstrating that our proposed model outperforms existing real-time OVD methods.
\end{itemize}

\section{Related Works}
\subsection{Open-Vocabulary Object Detection}
OVD signifies a crucial evolution in computer vision, marking a departure from conventional closed-set paradigms toward a more flexible open-vocabulary framework. Pioneering research in this domain \cite{9577418} was largely propelled by breakthroughs in vision-language models. These foundational methods employed pre-trained models, such as CLIP \cite{clip_ICML}, to empower detectors with the ability to recognize object categories defined by free-form text during the inference phase. More contemporary approaches reframe OVD as an image-text matching task, leveraging large-scale image-text corpora to substantially broaden the model's vocabulary. For instance, Grounding DINO \cite{liu2023grounding} integrates grounded pre-training with detection transformers through cross-modal fusion mechanisms. However, a significant drawback of these models is their substantial computational demand and architectural complexity, which precludes their use in real-time applications. In stark contrast, YOLO-World \cite{Cheng2024YOLOWorld} is specifically engineered to address this challenge. By achieving a dramatic improvement in inference speed over traditional two-stage OVD methods, it effectively meets the latency requirements of practical scenarios. This advancement is accelerating the transition of OVD technology from academic exploration to industrial deployment, unlocking novel potentials in domains such as intelligent surveillance, robotic vision, and autonomous driving.

\subsection{UAV-based Object Detection}
UAV-based object detection(UAV-OD) presents unique challenges, particularly in detecting small objects and managing occlusions. The prevailing paradigms for UAV-OD fall into two categories: two-stage and single-stage. While two-stage detectors generally yield superior accuracy due to their more complex pipelines, their practical application is often limited by the need for deployment on resource-constrained airborne platforms. Therefore, single-stage methods have gained greater traction in the research community, as they provide a more favorable trade-off between sufficient accuracy and the high inference speeds required for real-time operation. Key research directions for improving single-stage detectors primarily involve leveraging finer-grained feature maps (e.g., P2)\cite{11010850} for small object detection through multi-scale fusion\cite{TII1, chenYOLOMSRethinkingMultiScale2025}, formulating novel loss functions\cite{8237586}, and adopting high-resolution processing strategies such as image tiling with subsequent bounding box merging\cite{ESOD}.

\section{Datsets}
Current object detection datasets captured from drone perspectives are insufficiently large in scale and lack diversity in object categories, making them unable to support the pretraining of OVD models. To address this limitation, this paper introduces the UAVDE-2M and UAVCAP-15K datasets, which fulfill the pretraining requirements for UAV-OVD. 

\subsection{UAVDE-2M}
To construct an ultra-large-scale UAV-based OVD dataset, we introduce our UAV-Label Engine, by refining the existing LAE-Label Engine. The LAE-Label Engine simplistically categorizes existing remote sensing datasets into two types: Fully-annotated Datasets (FOD) and Completely Unlabeled Datasets (COD). For FOD, the engine directly slices the large-scale source images and then standardizes their format. For COD, it follows a sequential pipeline: it first employs the Segment Anything Model (SAM) to extract masks, then crops images based on these masks, uses a Large Vision-Language Model (LVLM) to predict categories, and finally performs class filtering.\par

Building upon the LAE-Label Engine, our UAV-Label Engine introduces a more sophisticated data handling pipeline. Its core innovation is the classification of data into a new category named Partially-annotated Datasets (FOP) which alongside the standard FOD and COD. This FOP category is specifically designed to address datasets with valuable but not immediately usable annotations, tackling common challenges such as: 1) frame redundancy in UAV video data, 2) incompatible annotation types (e.g., segmentation vs. rotated boxes), and 3) ambiguous class labels (e.g., `vehicle,' `obstacle').\par

To address these challenges, we implement a targeted three-stage pipeline. For redundant images, we first extract deep features from the image sequence using DINOv2 and then compute the cosine similarity between adjacent frames. If the similarity exceeds a predefined threshold, the redundant frame is discarded. For inconsistent annotation formats, we develop custom scripts to convert all annotations into a standardized bounding box format. Finally, to resolve class ambiguity, we crop the object instances corresponding to ambiguous labels and perform a fine-grained re-classification to assign more specific categories via LVLM.\par
	
\begin{figure}[htbp]
	\centering
	\subfloat[VisDrone]
	{\includegraphics[width=0.23\textwidth]{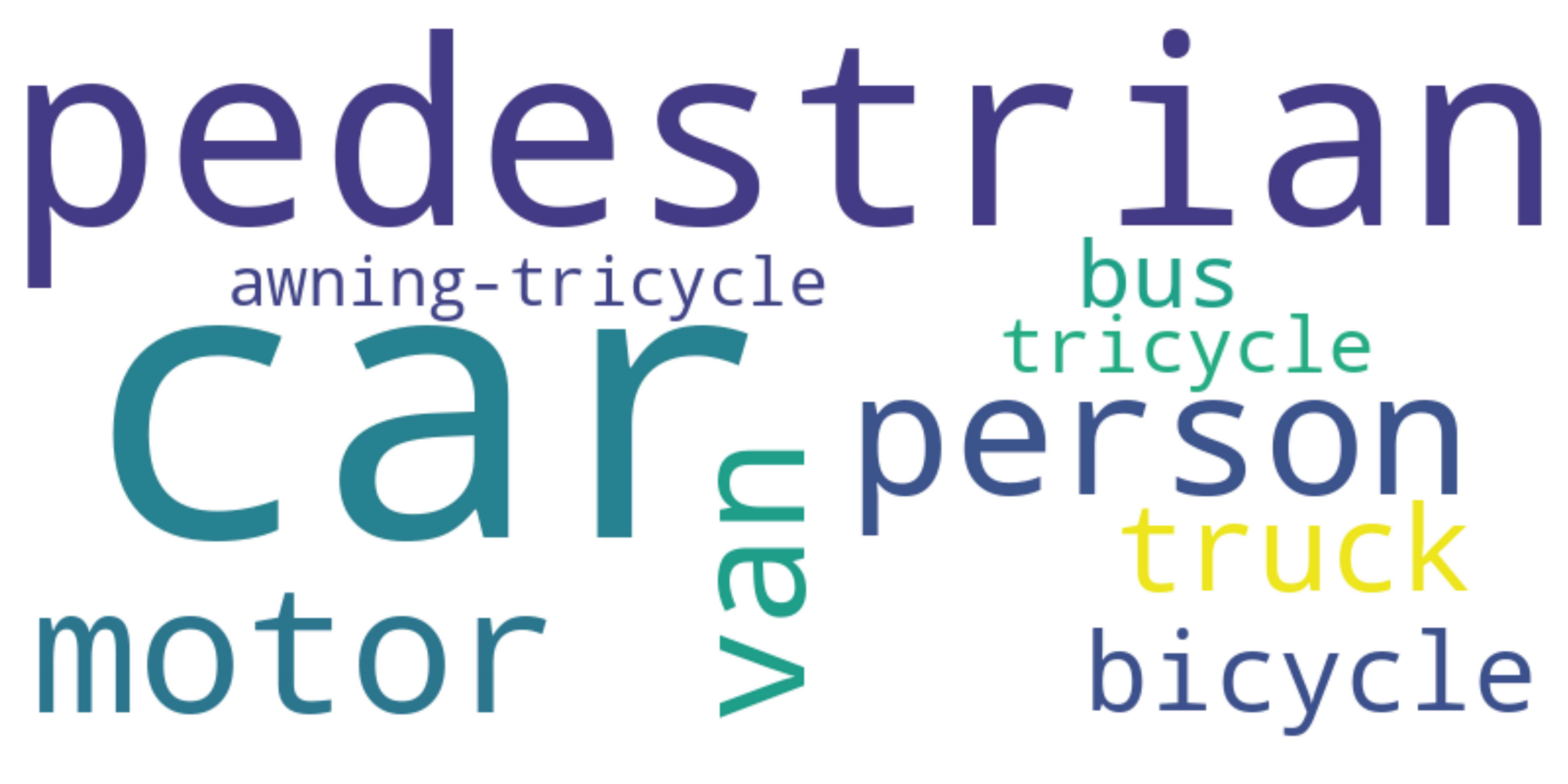}}\hfill
	\subfloat[CODrone]
	{\includegraphics[width=0.23\textwidth]{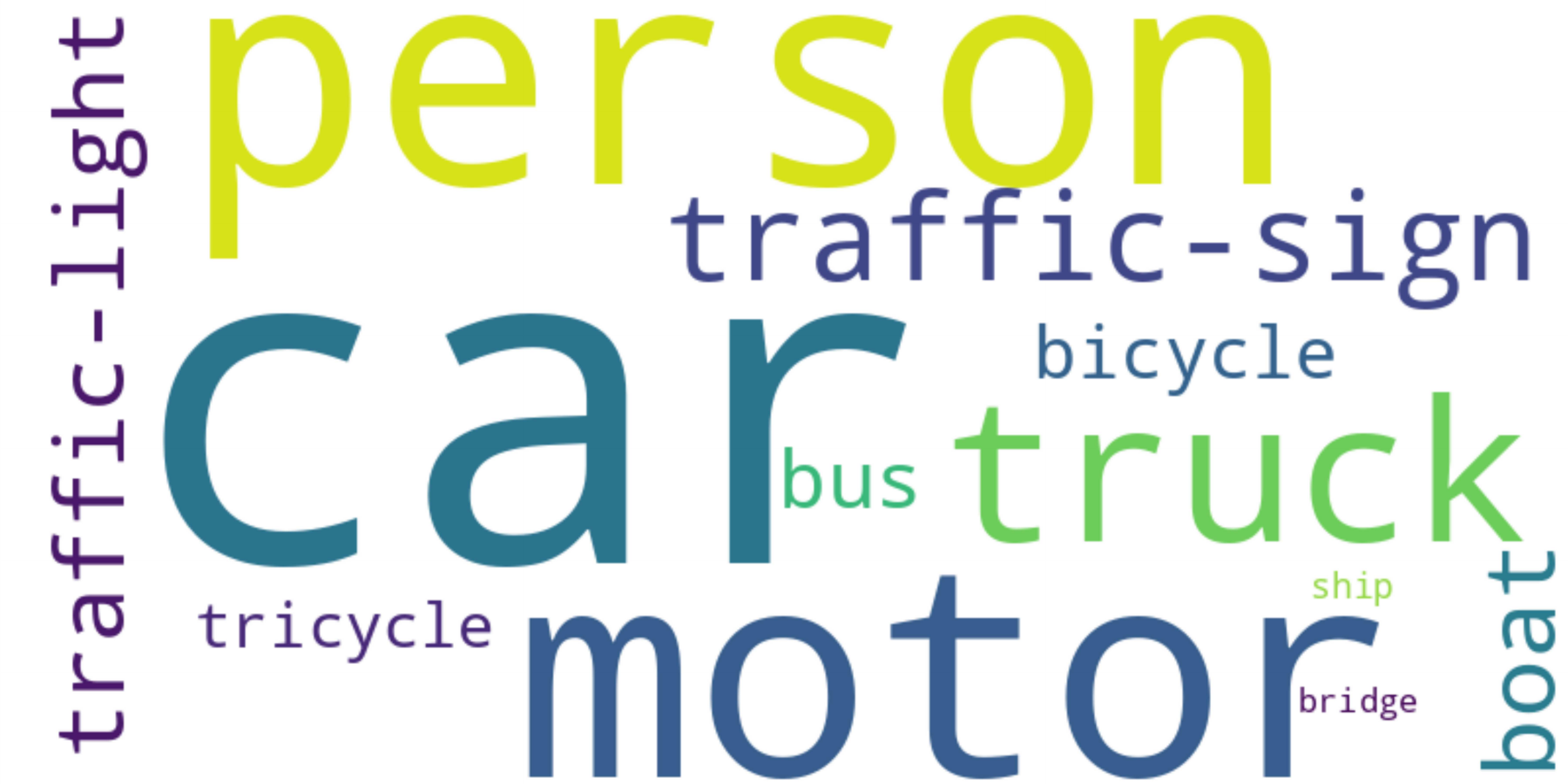}}\\
	\subfloat[ERA]
	{\includegraphics[width=0.23\textwidth]{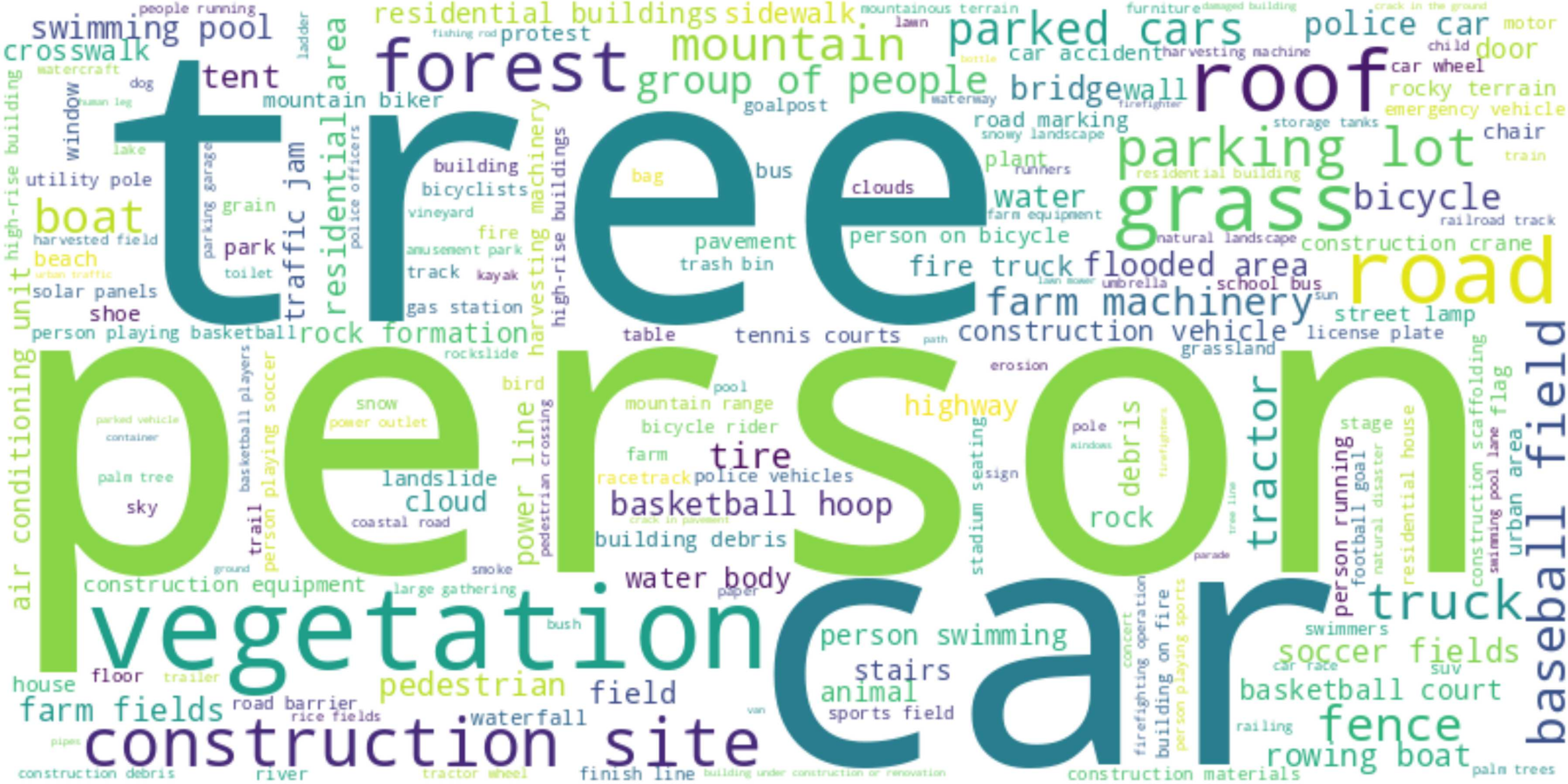}}\hfill
	\subfloat[FloodNet]
	{\includegraphics[width=0.23\textwidth]{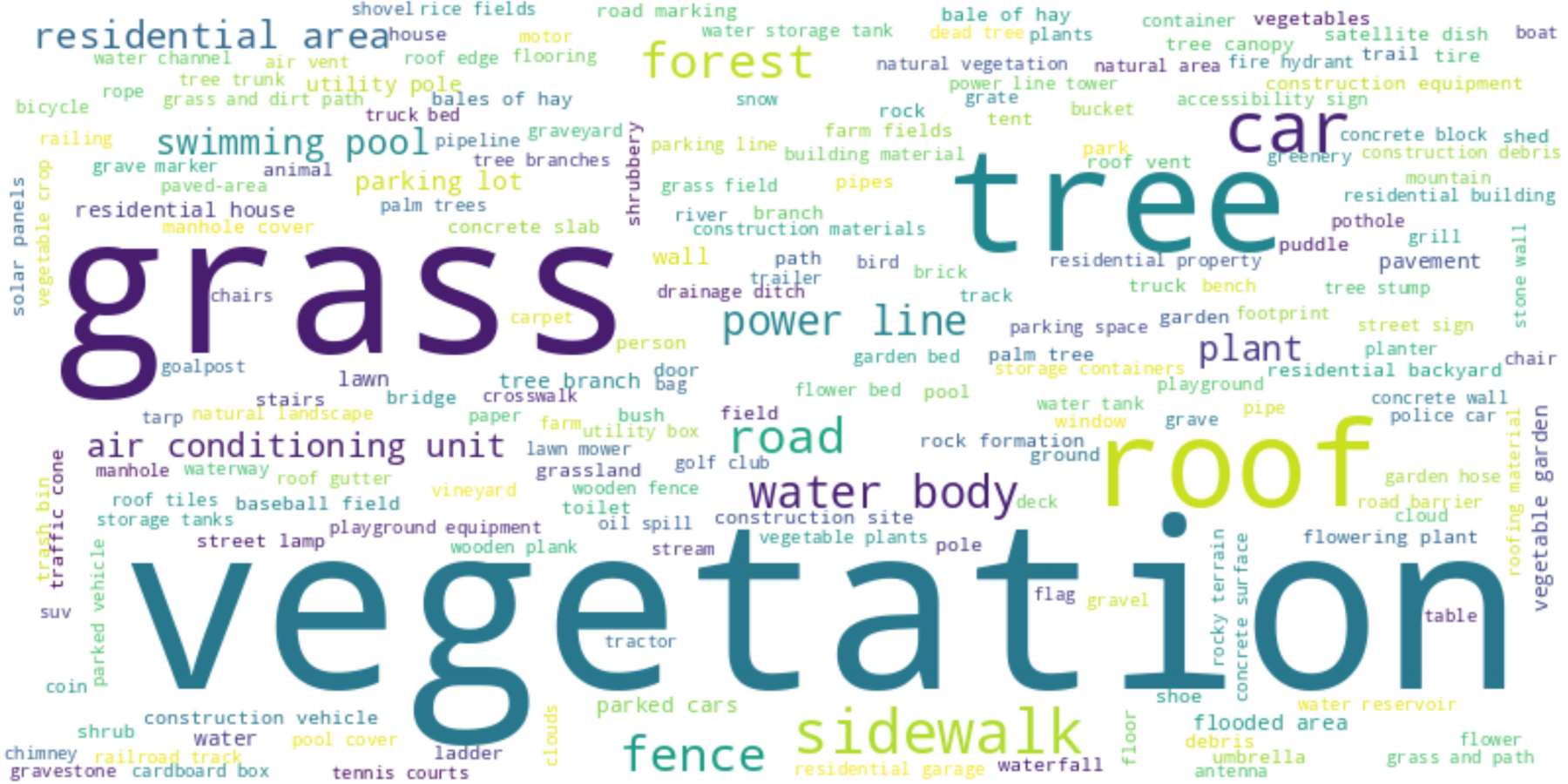}}\\
	\caption{\textbf{Word clouds illustrating the class distribution of the datasets.} In contrast to the pre-annotated FOD datasets (e.g., (a), (b)), the COD datasets processed by our LVLM pipeline (e.g., (c), (d)) exhibit a significantly richer and more diverse set of class labels.}
	\label{fig:word_clouds}
\end{figure}

The UAVDE-2M dataset we proposed is composed of multiple collected datasets processed using the UAV-Label Engine. The details of each sub-dataset constituting UAVDE-2M are shown in Table \ref{tab:detail of UAVDE-2M}. Word clouds illustrating the class distribution of the datasets shown in Figure \ref{fig:word_clouds}. Our UAVDE-2M dataset covers over 2.4 million instances, more than 1800 categories, and a total of about 245K images. Moreover, UAVDE-2M includes not only ground targets from a UAV perspective but also water surface objects (e.g. SeaDronesSee\cite{Seadronessee}, AFO\cite{AFO}) and aerial objects (e.g. detfly\cite{Det-Fly}).\par

\begin{table}[htbp]
	\centering
	\caption{\textbf{Detailed composition of the UAVDE-2M dataset.} Our UAVDE-2M dataset was constructed by sourcing and processing data from a diverse range of existing open-source datasets.}
	\label{tab:detail of UAVDE-2M}
	\begin{tabular}{cccc}
		\toprule[1pt]
		Datasets              & Images & Instances & Categories \\
		\hline
		\multicolumn{4}{c}{\textbf{COF}}                        \\
		\hline
		CARPK\cite{CARPK}                 	& 1448   & 89642     & 1          \\
		Cattle\cite{cattle}                	& 958    & 2079      & 1          \\
		LADD\cite{LADD}                  	& 3112   & 5019      & 1          \\
		SeaDronesSee\cite{Seadronessee}     & 27042  & 60566     & 5          \\
		UAVASTE\cite{uav-vast}              & 2432   & 3914      & 1          \\
		VisDrone\cite{visdrone}             & 8629   & 457066    & 10         \\
		Spanish Traffic\cite{xibany}        & 14490  & 152594    & 3          \\
		CODrone\cite{codrone}               & 50414  & 413750    & 12         \\
		DetFly\cite{Det-Fly}                & 10449  & 10449     & 1           \\
		\hline
		\multicolumn{4}{c}{\textbf{FOP}}                        \\
		\hline
		AeroScapes\cite{AeroScapes}         & 1110   & 8145      & 168        \\
		AFO\cite{AFO}                   	& 1811   & 7128      & 6          \\
		auair\cite{auair}                 	& 1190   & 4666      & 8          \\
		DAC-SDC\cite{DAC-SDC}              	& 4547   & 4547      & 12         \\
		RescueNet\cite{RescueNet}           & 20191  & 25500     & 88         \\
		uav123\cite{uav123}                	& 2091   & 12729     & 3          \\
		UAVDT-MOT\cite{uavdet}             	& 637    & 10080     & 3          \\
		UAVOD10\cite{UAVOD10}               & 844    & 18334     & 10         \\
		UUD\cite{UUD}                   	& 1164   & 20846     & 141        \\
		DroneVehicle\cite{DenseUAV} 		& 27654  & 451618    & 5          \\
		ICG Drone Dataset     				& 5121   & 35979     & 548        \\
		\hline
		\multicolumn{4}{c}{\textbf{COD}}                        \\
		\hline
		DenseUAV\cite{DenseUAV}             & 9096   & 156611    & 748        \\
		ERA\cite{era}                   	& 2790   & 32879     & 838        \\
		FLAME3\cite{FLAME3}                	& 3766   & 51323     & 184        \\
		FloodNet\cite{flootnet}             & 11585  & 285180    & 772        \\
		SUES\cite{SUES}                  	& 10829  & 32500     & 563        \\
		UAVDT-SOT\cite{uavdet}             & 1061   & 9029      & 304        \\
		University-1652\cite{University-1652}       & 21526  & 93282     & 643        \\
		\hline
		\multicolumn{4}{c}{\textbf{Images}: $\approx 245K$, \textbf{Instance}: $\approx 2.4M$, \textbf{Categories}: 1,853} \\
		\bottomrule[1pt]
	\end{tabular}
\end{table}

\subsection{UAVCAP-15K}
\begin{figure*}[htbp]
	\centering
	\includegraphics[width=\textwidth]{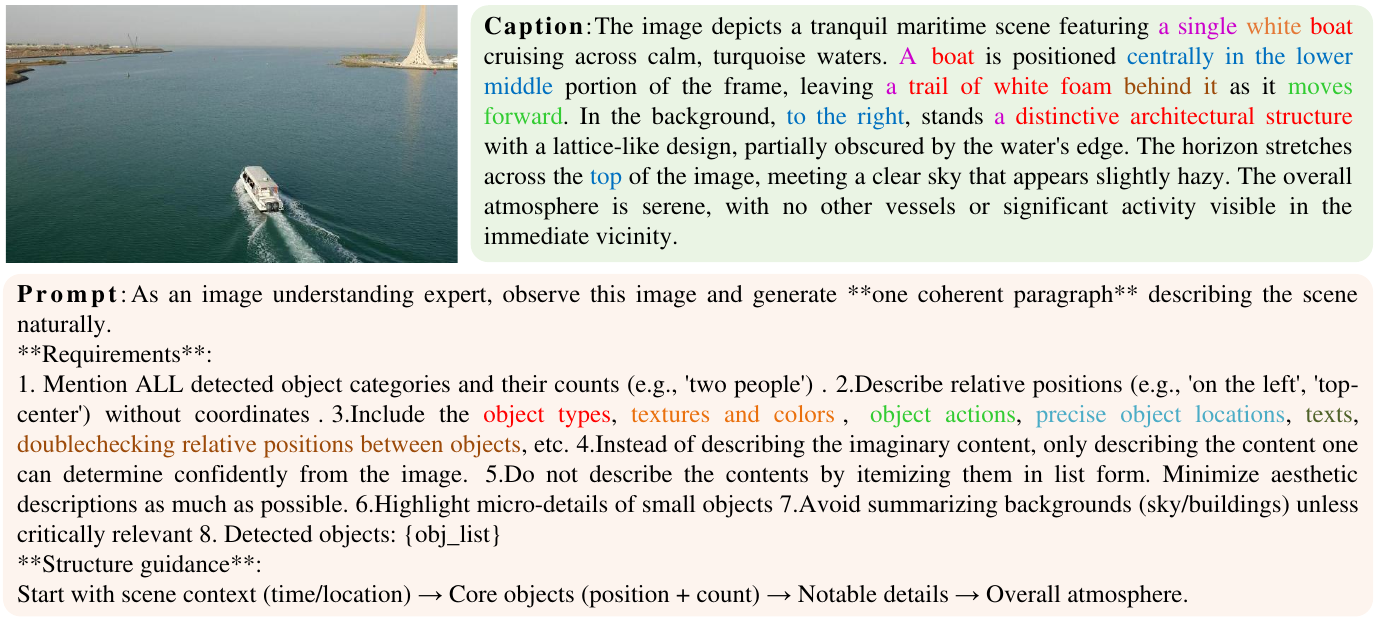}
	\caption{\textbf{Prompt and a sample of UAVCAP-15K.} The UAVCAP-15K dataset utilizes a format of long-form, detailed descriptions for its captions. Moreover, it employs prompting techniques to enforce a consistent structure, ensuring all captions describe the image content in a uniform manner.}
	\label{fig:UAVCAP-15k sample}
\end{figure*}

Compared to detection datasets, captioning datasets contain richer semantic information and are commonly used in conjunction with detection datasets during open-vocabulary detection (OVD) model pretraining. Datasets with detailed image captions significantly enhance the training of OVD models\cite{Fu_2025_CVPR}. To construct a large-scale UAV-based caption dataset, we selected suitable images from the UAVDE-2M dataset and leveraged LVLM to generate image captions by integrating existing annotations. Specifically, we convert the absolute coordinates of objects in images into relative positional descriptions (e.g., top-left corner, bottom-right corner) and integrate them as part of the prompt. Simultaneously, the LVLM is required to maximize the description of object types, object actions, precise object locations, texts, doublechecking relative positions between objects, etc\cite{Fu_2025_CVPR}. Given the prevalence of small-sized objects in UAV perspective images, our prompts explicitly emphasize the detailed description of small-scale objects while deliberately minimizing the description of large-scale background elements. We used the Qwen2.5-VL\cite{Qwen2.5-VL} as the LVLM to construct UAVCAP-15K. The prompt and sample of UAVCAP-15K is shown in Figure \ref{fig:UAVCAP-15k sample}.

\section{Methods}
\subsection{Cross-Attention Gated Enhancement Module}
\begin{figure*}[htbp]
	\centering
	\includegraphics[width=\textwidth]{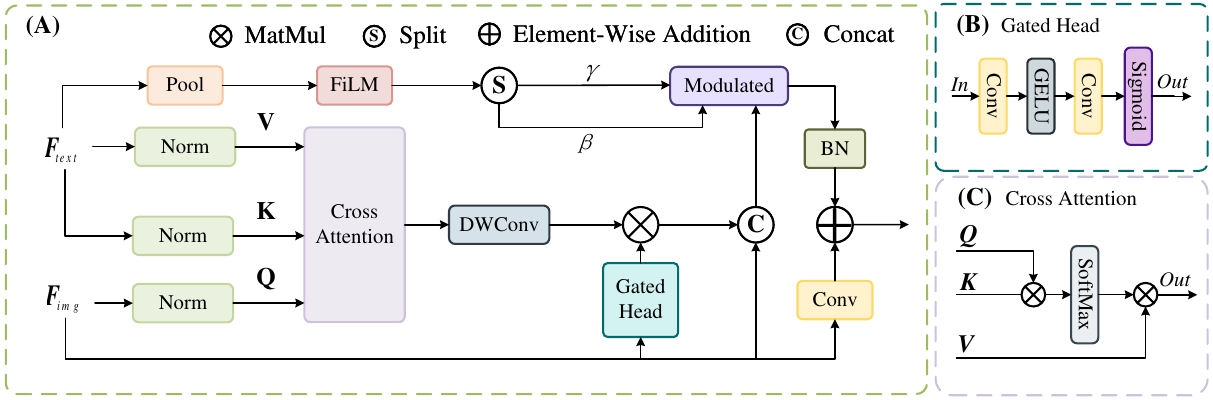}
	\caption{\textbf{Detail of cross-attention gated enhancement module.} The A is the overview of cross-attention gated enhancement module, B and C are the detail of gated head and cross attention respectively.}
	\label{fig:Fusion}
\end{figure*}

To effectively integrate natural language guidance with visual features, we propose a novel fusion module, termed CAGE (Context-Aware Gated Ensemble). It enriches spatial image features with semantic context derived from textual descriptions. The module operates on an image feature map $F_{img} \in \mathbb{R}^{B \times H \times W \times C}$ and a sequence of text embeddings $F_{text} \in \mathbb{R}^{B \times N \times D}$, where $D$ denotes batch size, $H, W$ are spatial dimensions, $C$ denotes channel depth, $D$ donates the number of categories, and $D$ donates the depth of the text embeddings. The module employs a multi-stage architecture combining: (1) a cross-modal attention mechanism, (2) a spatial context refinement block with an adaptive gate, (3) a global FiLM-based conditioning layer, and (4) a residual connection. The detailed structure of CAGE is shown in Figure \ref{fig:Fusion}.\par
The fusion process initiates by generating a spatial-semantic context map via multi-head cross-attention, where a query tensor, $Q$, is derived from the image features. The features are first projected by a $1 \times 1$ convolution, spatially flattened, and then normalized, a process which can be formulated as:
\begin{equation}
	Q = {LayerNorm}({Flatten}({Conv}_{1 \times 1}(F_{img})))
\end{equation}
where $P$ denotes the projection dimension. Currently, the key ($K$) and value ($V$) tensors are derived from the text features after layer normalization, followed by distinct linear transformations.
\begin{equation}
	F'_{text} = {LayerNorm}(F_{text})
\end{equation}

\begin{equation}
	\begin{aligned}
		K &= F'_{text} W_K \in \mathbb{R}^{B \times L \times P} \\
		V &= F'_{text} W_V \in \mathbb{R}^{B \times L \times P}
	\end{aligned}
\end{equation}

where $W_K$ and $W_V$ are the weight matrices of the linear projection layers. The attention scores are computed via scaled dot-product attention across $h$ heads, each with dimension $d_k = P/h$. For the $i$-th head, the operation is:
\begin{equation}
	head_i = {softmax}\left(\frac{Q_i K_i^T}{\sqrt{d_k}}\right)V_i
\end{equation}
The outputs of all heads are concatenated and reshaped to form the context map ${ctx}_{map} \in \mathbb{R}^{B \times P \times H \times W}$.\par

In gated context refinement, the raw context map $ctx_{map}$ undergoes a refinement and gating process to enhance local features. The context map is first projected to an intermediate channel dimension and then refined by a block $\mathcal{F}_{dw}$ consisting of depthwise separable convolutions and GELU activations. An optional occlusion gate generates a per-pixel probability map $G \in [0, 1]^{B \times 1 \times H \times W}$ from the input image features, which determines the relevance of the textual context at each location, formulated as:
\begin{equation}
	G = \sigma(Conv_{1 \times 1}(GELU(Conv_{3 \times 3}(F_{img}))))
\end{equation}
where $\sigma$ is the Sigmoid function. The refined context is then modulated by this gate:
\begin{equation}
	ctx_{\text{ref}} = G \odot \mathcal{F}_{dw}(Conv_{1 \times 1}(ctx_{map}))
\end{equation}
where $\odot$ denotes element-wise multiplication. \par
The final stage integrates the refined context with the original image features using both concatenation and global modulation. The gated context $	ext{ctx}_{	ext{ref}}$ is concatenated with the original image features $F_{img}$ along the channel dimension and subsequently fused via a convolutional block $\mathcal{F}_{merge}$ to produce a pre-modulated feature map:
\begin{equation}
	F_{pre-film} = \mathcal{F}_{merge}([F_{img}; 	ctx_{ref}])
\end{equation}
For global semantic control, a FiLM layer is utilized. The text features $F_{text}$ are globally pooled to a sentence-level vector $f_{pool}$. This vector is passed through a multi-layer perceptron $\mathcal{F}_{film}$ to predict the channel-wise affine transformation parameters $\gamma$ and $\beta$:
\begin{equation}
	[\gamma, \beta] = \mathcal{F}_{film}(f_{pool}) \quad (\gamma, \beta \in \mathbb{R}^{B \times C_{out} 	\times 1 \times 1})
\end{equation}
The pre-modulated features are then transformed as follows:
\begin{equation}
	F_{modulated} = (1 + \gamma) \odot F_{pre-film} + \beta
\end{equation}
This operation allows the text command to scale and shift the visual feature activations dynamically.\par
To preserve the original visual information and ensure stable gradient flow, the module incorporates a parallel residual path, $F_{res} = \mathcal{F}_{res}(F_{img})$, where $\mathcal{F}_{res}$ is either a $1 	\times 1$ convolution or an identity map. The final output $F_{out}$ is the sum of the residual path and the batch-normalized modulated features:
\begin{equation}
	F_{out} = F_{res} + BatchNorm(F_{modulated})
\end{equation}
This design ensures that the fusion process enriches the visual features without catastrophically altering them, leading to a robust and effective fusion.\par

\subsection{Module Integration Strategy}
To seamlessly integrate the proposed CAGE module, we replace the T-CSPLayer blocks in the YOLO-World-v2 neck with our design. Each CAGE module receives multi-scale visual features from the PAN structure and text embeddings from the shared text encoder. The output dimensions are carefully aligned with those of the replaced T-CSPLayer, ensuring that the downstream architecture remains unchanged. This plug-and-play design allows CAGE to serve as a direct substitute for existing text–vision fusion modules without introducing additional architectural overhead.

\subsection{Rationale and Advantages}
Compared with the original T-CSPLayer, CAGE provides a more fine-grained and robust text–vision alignment. Its multi-head cross-attention enhances local semantic grounding, the occlusion-aware gating mechanism adaptively filters irrelevant context, and the global FiLM pathway modulates features at the channel level. These enhancements collectively yield richer cross-modal representations while preserving essential visual details through residual connections. By adopting a drop-in replacement strategy, CAGE not only improves detection accuracy but also demonstrates strong generalizability, making it easily transferable to other vision–language frameworks.

\section{Experiments}
\subsection{Implementation details}
\paragraph{Data.}
Following prior work\cite{wang2025yoloerealtimeseeing,Cheng2024YOLOWorld}, we jointly pre-train our model on a combination of detection and captioning datasets, namely UAVDE-2M and UAVCAP-15K, where images from VisDrone\cite{visdrone} are excluded.\par
\paragraph{Training.}
All models we trained were processed for 100 epochs on a setup of 4$\times$A800 GPUs. We set the batch-size as 128, for data augmentation, we adopted the default strategy and parameters provided by Ultralytics\cite{ultralytics}.\par
\paragraph{Evaluation.}
During the evaluation phase, we employ a text-prompting approach. For each dataset, we utilize its class names directly as the text prompts.

\subsection{Comparison Results} % Or \subsection{...} depending on your structure
The zero-shot evaluation results of our proposed method on the VisDrone test set are presented in Table~\ref{tab:results visdrone}. The analysis first highlights the efficacy of domain-specific pre-training. Compared to the official YOLO-World-v2 models pre-trained on a general-purpose dataset (OGC), the models trained on our custom UAVDE-2M and UAVCAP-15K datasets exhibit substantial performance improvements. For instance, the mAP of the YOLO-World-v2-L model significantly increases from 8.59 to 12.2, which strongly demonstrates that our datasets can effectively enhance the model's zero-shot generalization capability in aerial imagery scenarios. Furthermore, our proposed method demonstrates consistent superiority over the baselines under identical training conditions. Our models, outperform the original YOLO-World-v2 across all scales (S/M/L), achieving mAP scores of 10.0, 12.5, and 13.9, respectively, which are notable improvements over the baseline scores of 9.6, 11.1, and 12.2. In addition to its accuracy advantages, our method also excels in model efficiency. Taking the L-scale model as an example, ours boosts the mAP from 12.2 to 13.9 while simultaneously reducing the parameter count from 48M to 34M and GFLOPs from 204.5 to 144.0. This result underscores the ability of our method to significantly reduce model complexity and computational cost without sacrificing performance, making it highly suitable for resource-constrained edge computing applications. In summary, the experimental results thoroughly validate the dual effectiveness of our proposed datasets and method, achieving concurrent improvements in both accuracy and efficiency for the task of zero-shot object detection in UAV-based images.\par
\begin{table*}[htbp]
	\centering
	\caption{\textbf{Zero-shot Evaluation on VisDrone.} We evaluate out methods on VisDrone\cite{visdrone} test set in a zero-shot manner. For models trained on standard benchmark datasets, we adopt the weights released by Ultralytics\cite{ultralytics}. OGC indicates Objects365\cite{shao2019objects365}, GoldG\cite{GoldG} and CC3M\cite{Cheng2024YOLOWorld}. UAV indicates to UAVDE-2M and UAVCAP-15K.}
	\label{tab:results visdrone}
	\begin{tabular}{cccccccc}
		\toprule[1pt]
		Methods               			& Data & Size & Epochs 					 	& Params(M) & GFLOPs & AP50 & mAP \\
		\hline
		YOLO-World-S          		& OGC       & 640  & -                          & 13     	& 71.5   & 7.44  & 4.32 \\
		YOLO-World-M          		& OGC       & 640  & -                          & 29     	& 131.4  & 10.9  & 6.64 \\
		YOLO-World-L          		& OGC       & 640  & -                          & 48     	& 225.6  & 12.9  & 8.00 \\
		YOLO-World-v2-S       		& OGC       & 640  & -                          & 13     	& 51.0   & 7.87  & 4.58 \\
		YOLO-World-v2-M       		& OGC       & 640  & -                          & 29     	& 110.5  & 11.1  & 6.83 \\
		YOLO-World-v2-L       		& OGC       & 640  & -                          & 48     	& 204.5  & 13.9  & 8.59 \\
		YOLO-World-v2-S       		& UAV       & 640  & 100                        & 13     	& 51.0   & 16.0  & 9.6	\\ \rowcolor{gray!30}
		YOLO-World-v2-S w/ CAGE 	& UAV       & 640  & 100                        & \textbf{11}     	& \textbf{46.6}   & \textbf{16.4}  & \textbf{10.0} \\
		YOLO-World-v2-M       		& UAV       & 640  & 100                        & 29     	& 110.5  & 17.7  & 11.1 \\ \rowcolor{gray!30}
		YOLO-World-v2-M w/ CAGE 	& UAV       & 640  & 100                        & \textbf{22}     	& \textbf{84.2}   & \textbf{19.4}  & \textbf{12.5} \\ 
		YOLO-World-v2-L       		& UAV       & 640  & 100                        & 48     	& 204.5  & 19.4  & 12.2 \\ \rowcolor{gray!30}
		YOLO-World-v2-L w/ CAGE 	& UAV       & 640  & 100                        & \textbf{34}     	& \textbf{144.0}  & \textbf{21.2}  & \textbf{13.9} \\ 
		\bottomrule[1pt]
	\end{tabular}
\end{table*}
To further validate the generalization capability of our curated datasets, we conducted a cross-domain zero-shot evaluation. Specifically, models pre-trained on our UAVDE-2M and UAVCAP-15K datasets were tested on the SIMD remote sensing dataset~\cite{SIMD}, whose image characteristics are related yet distinct from our UAV-based images. The results are presented in Table~\ref{tab:results SIMD}. The outcomes unequivocally demonstrate that pre-training on our datasets significantly enhances the model's zero-shot performance in a novel domain. Compared to models trained on the general-purpose OGC dataset, the mAP of the YOLO-World-v2-L model trained on our datasets surged from 8.93 to 11.2, a substantial increase of 25.4\%. This finding strongly suggests that our UAVDE-2M and UAVCAP-15K datasets enable the model to learn more generalizable visual representations that are not confined to UAV-specific scenarios but can be effectively transferred to broader remote sensing tasks. Furthermore, our proposed models also exhibit strong competitive performance in this cross-domain challenge. Our model surpasses its baseline with an mAP of 10.56, while our L-scale model achieves a highly competitive performance comparable to its baseline. In conclusion, this experiment fully substantiates the generalization value of our datasets, confirming that they equip the model with robust knowledge that transcends specific domain boundaries.\par
\begin{table}[htbp]
	\centering
	\caption{\textbf{Zero-shot Evaluation on SIMD.} We evaluate out methods on SIMD\cite{SIMD} test set, which is a remote sensing dataset different with uav-based images.}
	\label{tab:results SIMD}
	\begin{tabular}{cccc}
		\toprule[1pt]
		Methods               		& Data 				& AP50  & mAP   \\
		\hline
		YOLO-World-S          		& OGC               & 7.18  & 4.24  \\
		YOLO-World-M          		& OGC               & 9.98  & 5.79  \\
		YOLO-World-L          		& OGC               & 14.64 & 9.20  \\
		YOLO-World-v2-S       		& OGC               & 8.61  & 4.75  \\
		YOLO-World-v2-M       		& OGC               & 12.07 & 7.24  \\
		YOLO-World-v2-L       		& OGC               & 15.03 & 8.93  \\
		YOLO-World-v2-M       		& UAV       		& 14.7  & 9.61  \\ \rowcolor{gray!30}
		YOLO-World-v2-M w/ CAGE 	& UAV       		& \textbf{15.9}  & \textbf{10.56} \\
		YOLO-World-v2-L       		& UAV       		& 16.9  & 11.2  \\ \rowcolor{gray!30}
		YOLO-World-v2-L w/ CAGE 	& UAV       		& \textbf{18.2}  & \textbf{11.0} \\
		\bottomrule[1pt]
	\end{tabular}
\end{table}
To dissect the specific contributions of our proposed datasets to the model's performance, we conducted a series of ablation studies in VisDrone\cite{visdrone} test set, with the results detailed in Table \ref{tab:Ablation data}. We established a YOLO-World-v2-S using a model pre-trained solely on the general-purpose OGC dataset, which achieved an mAP of only 4.58.\par
When we exclusively used our UAVDE-2M dataset for pre-training, the model's performance experienced a substantial leap, with the mAP soaring to 9.22. This dramatic improvement unequivocally demonstrates that the UAVDE-2M dataset provides the essential domain-specific knowledge required for effective zero-shot detection in UAV scenarios. Subsequently, by incorporating the UAVCAP-15K dataset for joint pre-training, the mAP was further boosted to 9.60. This indicates that UAVCAP-15K offers complementary and valuable information to UAVDE-2M, and their combination produces a synergistic effect that further enhances the model's detection capabilities. In summary, this ablation study clearly validates our data curation strategy, confirming that UAVDE-2M serves as the foundational contributor to the performance gains, while UAVCAP-15K acts as a beneficial supplement, together yielding the optimal performance.
\begin{table}[htbp]
	\centering
	\caption{\textbf{Ablation Study of pre-trained datasets.} We performed ablation experiments on the VisDrone\cite{visdrone} test set with different pre-trained data.}
	\label{tab:Ablation data}
	\begin{tabular}{ccccc}
		\toprule[1pt]
		OGC & UAVDE-2M & UAVCAP-15K & AP50 & mAP  \\
		\hline
		\checkmark  &       &        & 7.87 & 4.58 \\
		& \checkmark     &        & 15.5 & 9.22 \\
		& \checkmark     & \checkmark      & 16.0   & 9.60 \\
		\bottomrule[1pt]
	\end{tabular}
\end{table}
\subsection{Real World Application}

\begin{figure}[htbp]
	\centering
	\includegraphics[width=0.45\textwidth]{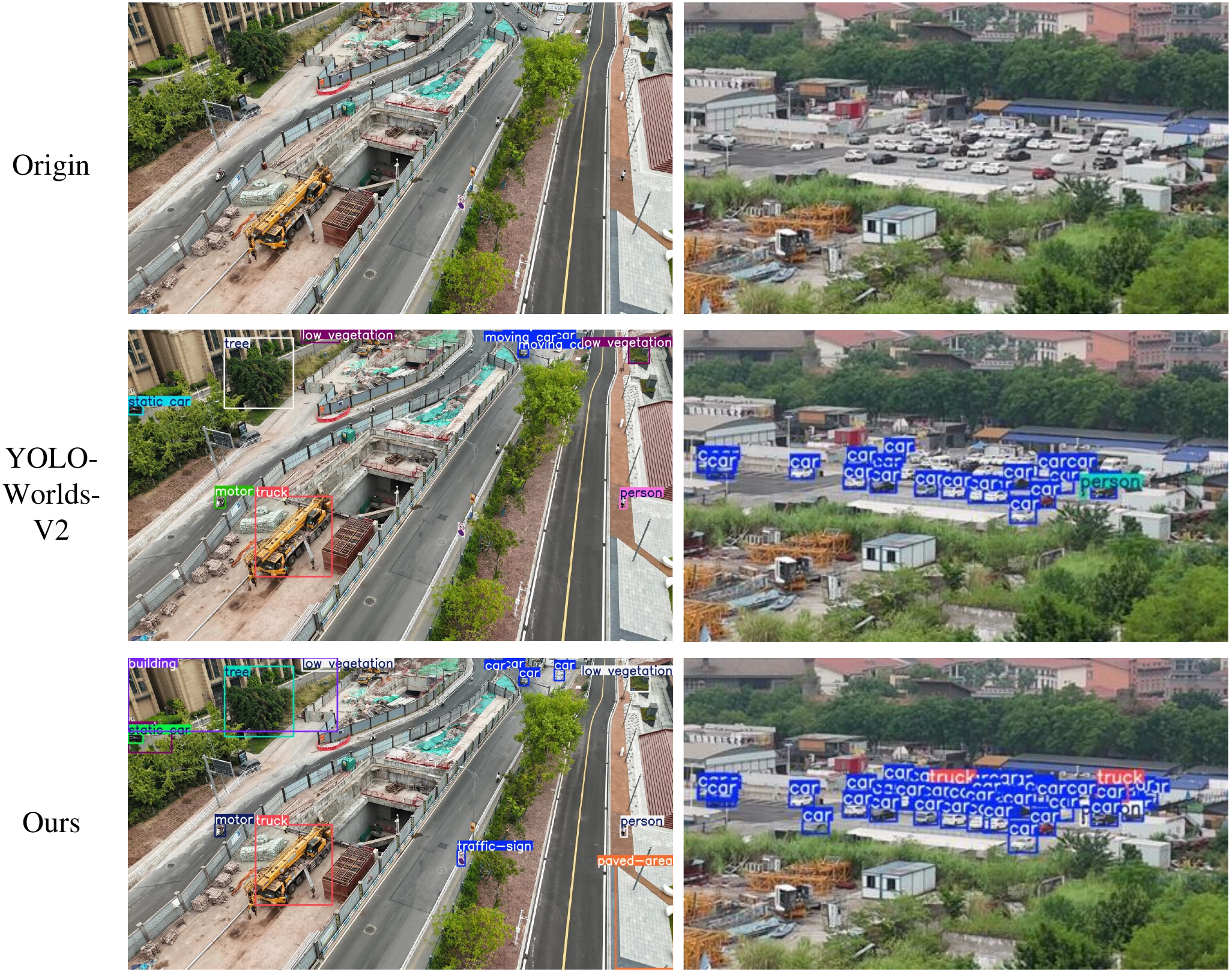}
	\caption{\textbf{Visualization of open-vocabulary detection results from a real-world UAV perspective.} Our method demonstrates better detection performance.}
	\label{fig:real world results}
\end{figure}

To validate the performance of our proposed model on the real-world UAV-based open-vocabulary object detection task, as well as its real-time operational capabilities on edge computing devices, we deployed our proposed model on an NVIDIA Jetson Orin NX 16G embedded computing device. This device was mounted on a P600 UAV, enabling the system to capture the video feed from the UAV's gimbal camera, perform inference in real-time, and subsequently transmit the detection results back to a ground station via a video transmission module. Table \ref{tab:orinnx infer} presents a comparison of the inference speeds between our proposed model and the baseline models on the NVIDIA Jetson Orin NX 16G. The results are shown in Figure \ref{fig:real world results}. Our proposed method outperforms YOLO-World in terms of performance, as highlighted in the second column, our method demonstrates superior performance over YOLO-World-v2, particularly in challenging scenes with dense objects and heavy occlusion(e.g. parking lots).
\begin{table}[]
	\centering
	\caption{\textbf{Comparison of Inference Speeds on the NVIDIA Jetson Orin NX.} TensorRT models are benchmarked with FP16 precision, ONNX models are using CPU inference.}
	\label{tab:orinnx infer}
	\begin{tabular}{ccc}
		\toprule[1pt]
		\diagbox{Methods}{Type}& TensorRT(ms) 				& ONNX(ms)  \\ \hline
		YOLO8s-WorldV2 			& 25.38       				& 527 	\\
		YOLOE-11s 				& 28.34       				& 591 	\\
		YOLO8s-WorldV2 w/ CAGE	& \textbf{22.90}    		& \textbf{477}  	\\
		\bottomrule[1pt]
	\end{tabular}
\end{table}

\section{Conclusion}

\iffalse
We present UAVDE-2M and UAVCAP-15K, the first large-scale UAV-oriented OVD datasets, and CAGE, a lightweight gated cross-attention module. Together they push VisDrone zero-shot mAP from 8.59 to 13.9 while cutting 29 \% params and 30 \% FLOPs. Cross-domain tests on SIMD show +25 \% mAP, and real-time deployment on Jetson Orin NX runs at 44 FPS. Our data and plug-in module enable robust, real-time open-vocabulary detection from UAVs.
\fi

In this work we address the long-standing domain gap that handicaps open-vocabulary object detection (OVD) once the viewpoint moves from ground to drone. First, we introduce UAVDE-2M and UAVCAP-15K currently the largest and most diverse UAV-specific datasets—covering 2.4 M instances across 1 853 categories and 15K richly-captioned images. A refined UAV-Label Engine automatically resolves redundancy, annotation inconsistency and class ambiguity, providing clean yet exhaustive annotations in hours rather than months. Second, we propose CAGE, a dual-path cross-attention gated enhancement module that fuses language and vision with a controllable gate, global FiLM modulation and residual bypass. Plugged into YOLO-World-v2 with no other changes, CAGE yields:
\begin{itemize}
	\item +5.3 mAP on VisDrone zero-shot (8.59 → 13.9 for L-scale)
	\item –29 \% parameters and –30 \% GFLOPs, enabling 22.9ms latency on NVIDIA Jetson Orin NX
	\item +25.4 \% mAP cross-domain on the SIMD remote-sensing benchmark
\end{itemize}
Ablation confirms that UAVDE-2M supplies the core domain knowledge and UAVCAP-15K provides complementary semantic supervision; together they are indispensable. Real-world flight tests further verify robust detection under varying altitude, illumination and weather. By open-sourcing the datasets, code and TensorRT models, we hope to accelerate the deployment of truly open-world, real-time perception systems on resource-constrained aerial platforms.
\clearpage
{
    \small
    \bibliographystyle{ieeenat_fullname}
    \bibliography{main}
}

\end{document}